\pdfoutput=1

\documentclass[10pt,twocolumn,letterpaper]{article}

\usepackage{graphicx}
\usepackage{tabularray}
\usepackage{multirow}
\usepackage{bbding}
\usepackage[pagenumbers]{cvpr} 
\usepackage{footmisc}  

\definecolor{cvprblue}{rgb}{0.21,0.49,0.74}
\usepackage[pagebackref,breaklinks,colorlinks,allcolors=cvprblue]{hyperref}

\def\paperID{12472} 
\def\confName{CVPR}
\def\confYear{2025}

\title{DAMM-Diffusion: Learning Divergence-Aware Multi-Modal Diffusion Model for Nanoparticles Distribution Prediction}

\author{
    Junjie Zhou$^{1,2}$, \; 
    Shouju Wang$^3$\footnotemark[1], \; 
    Yuxia Tang$^3$, \;
    Qi Zhu$^{1,2}$, \;
    Daoqiang Zhang$^{1,2}$\footnotemark[1], \
    Wei Shao$^{1,2}$\footnotemark[1] \\
    {$^1$The College of Artificial Intelligence, Nanjing University of Aeronautics and Astronautics}\\ 
    {$^2$The Key Laboratory of Brain-Machine Intelligence Technology, Ministry of Education}\\ 
    {$^3$Department of Radiology, Nanjing Medical University}\\
    {\tt\small junjiezhou@nuaa.edu.cn,~dqzhang@nuaa.edu.cn,~shaowei20022005@nuaa.edu.cn}
    \vspace{-2mm}
}

\begin{document}

\maketitle
\footnotetext[1]{Corresponding authors}

\begin{abstract}
The prediction of nanoparticles (NPs) distribution is crucial for the diagnosis and treatment of tumors. Recent studies indicate that the heterogeneity of tumor microenvironment (TME) highly affects the distribution of NPs across tumors. Hence, it has become a research hotspot to generate the NPs distribution by the aid of multi-modal TME components. However, the distribution divergence among multi-modal TME components may cause side effects i.e., the best uni-modal model may outperform the joint generative model. To address the above issues, we propose a \textbf{D}ivergence-\textbf{A}ware \textbf{M}ulti-\textbf{M}odal \textbf{Diffusion} model (i.e., \textbf{DAMM-Diffusion}) to adaptively generate the prediction results from uni-modal and multi-modal branches in a unified network. In detail, the uni-modal branch is composed of the U-Net architecture while the multi-modal branch extends it by introducing two novel fusion modules i.e., Multi-Modal Fusion Module (MMFM) and Uncertainty-Aware Fusion Module (UAFM). Specifically, the MMFM is proposed to fuse features from multiple modalities, while the UAFM module is introduced to learn the uncertainty map for cross-attention computation. Following the individual prediction results from each branch, the Divergence-Aware Multi-Modal Predictor (DAMMP) module is proposed to assess the consistency of multi-modal data with the uncertainty map, which determines whether the final prediction results come from  multi-modal or uni-modal predictions. We predict the NPs distribution given the TME components of tumor vessels and cell nuclei, and the experimental results show that DAMM-Diffusion can generate the distribution of NPs with higher accuracy than the comparing methods. Additional results on the multi-modal brain image synthesis task further validate the effectiveness of the proposed method. The code is released \footnote[2]{\href{https://github.com/JJ-ZHOU-Code/DAMM-Diffusion}{https://github.com/JJ-ZHOU-Code/DAMM-Diffusion}}.
\end{abstract}

\begin{figure}[htbp]
 \centering
 \includegraphics[width=\linewidth]{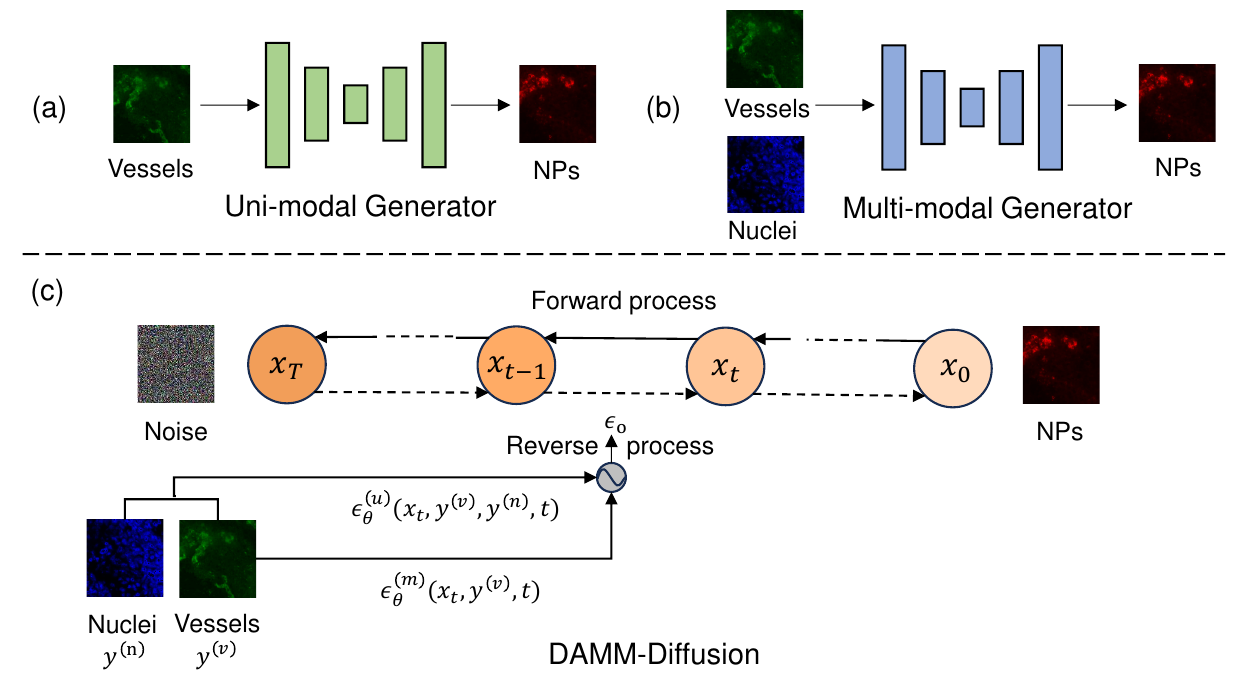}
 \caption{The illustration of different NPs distribution prediction methods. (a) The Uni-modal method predicts the distribution of NPs by vessels. (b) The multi-modal method predicts the distribution of NPs by the combination of vessels and nuclei. (c) Our DAMM-Diffusion considers both uni-modal and multi-modal branches for NPs distribution prediction by considering the divergence among nuclei and vessels channels.}
 \vspace{-3mm}
 \label{fig:fig1}
\end{figure}
\section{Introduction}
\vspace{-2mm}
Nanoparticles (NPs) present a promising avenue for the diagnosis and treatment of tumors~\cite{jin2020application,dessale2022nanotechnology}. Leveraging the Enhanced Permeability and Retention (EPR) effect,
NPs can selectively accumulate in tumor tissues due to their leaky blood vessels and compromised lymphatic drainage~\cite{fang2011epr,liao2020physical}. Hence, the accurate prediction of NPs distribution can help deliver drugs to the tumor site, resulting in improved therapeutic outcomes~\cite{shen2024enhancing}. Recent studies indicate that the tumor microenvironment (TME) components, such as blood vessels and nuclei, highly affect the delivery and distribution of NPs across tumors~\cite{goyal2023advances,shen2024enhancing}. It has become a research hotspot to infer the distribution of NPs by considering the spread of multiple TME components basing on the fluorescence~\cite{ganguly2023bioimaging} or mass spectrometry imaging~\cite{cazier2020development}. But until now, accurately predicting intratumoral nanomedicine distribution remains challenging due to the complex spatial relationship between different tumor microenvironments and nanomedicine accumulation~\cite{xu2023high}.

Recently, with the rapid development of deep learning technology, the generative models have already achieved significant success for image synthesis~\cite{li2019diamondgan,dalmaz2022resvit,jiang2023cola} and density estimation~\cite{singh2018nonparametric, liu2021density}. Leveraging the powerful modeling capabilities of generative models, we can establish the correlation between specific tumor components and NPs distribution. For example, vessels have long been regarded as highly relevant to NPs distribution~\cite{torosean2013nanoparticle,chauhan2011delivery,chauhan2020normalization,zhang2017captopril}, and thus Tang \emph{et al.}~\cite{tang2021ganda} firstly explore the deep generative models to predict the NPs distribution by the vessel channel. Besides the vessel channel, the stained DAPI images can provide the nuclei localization information of tumor cells that are also associated with the distribution of NPs \cite{sousa2010functionalized}. In view of that, recent researches have explored the application of multi-modal learning for predicting the distribution of NPs by the combination of both vessels (stained by CD31 marker) and nuclei (stained by DAPI)~\cite{tang2021ganda, xu2023high} channels. For example, Tang \emph{et al.}~\cite{tang2021ganda} devise a multi-modal generative adversarial network aimed at predicting NPs distribution conditioned with both tumor vessels and nuclei information. By utilizing the same dataset, Xu \emph{et al.}~\cite{xu2023high} present a high-resolution prediction network (HRPN) to generate pixel-level distribution of intratumoral NPs with superior performance. All these studies indicate that different biomarkers can provide complementary information for estimating the distribution of NPs. \\
\indent Although much progress has been achieved, the existing generative models are either hard to train \cite{tang2021ganda} or have difficulty in matching the distribution of real images with satisfied precision \cite{xu2023high}. In the past few years, the Diffusion models ~\cite{song2019generative,ho2020denoising,song2020score} have been developed as a powerful generative models that are capable of generating high-quality images with a more stable training process. In the scenario of multi-modal image generation, Feng \emph{et al.}~\cite{feng2023diffdp} propose a novel diffusion-based dose prediction (DiffDP) model for predicting the radiotherapy dose
by incorporating multi-modal anatomical information. Jiang \emph{et al.}~\cite{jiang2023cola} have proposed a conditioned latent diffusion model for multi-modal MRI image synthesis. Huang \emph{et al.}~\cite{huang2023collaborative} propose a Collaborative Diffusion model to generate face image with the multi-modal inputs. However, all these multi-modal diffusion models often neglect the inherent divergence~\cite{yang2022domfn} among the input multi-modal data. For the NPs distribution prediction problem, the divergence between nuclei and vessels may detrimentally impact results. In other words, although the nuclei are treated as an auxiliary modality for better generating the distribution of NPs ~\cite{tang2021ganda,xu2023high}, the best uni-modal model based on the vessel channel may even outperform the multi-modal generative model (shown in \cref{tab:table1} of \cref{sec:exp_comparison}). These observations inspire us to address the divergence among the input multi-modal data to predict the distribution of NPs. \\
\indent To address the above issues, we propose a Divergence-Aware Multi-Modal Diffusion model ({i.e.,} DAMM-Diffusion) for predicting the distribution of NPs by the combination of tumor vessels (stained by CD31 marker) and nuclei information (stained by DAPI marker). 
As shown in \cref{fig:fig1}, different from the previous uni-modal or multi-modal generators, DAMM-Diffusion considers both uni-modal and multi-modal branches to perform the reverse step of the diffusion model in a unified network. These two branches output the predictions individually with the U-Net network and a proposed Divergence-Aware Multi-Modal Predictor (DAMMP) decides the final output relied on multi-modal predictions or only using the uni-modal generation results. To evaluate the effectiveness of our methods for NPs distribution prediction, we conduct experiments on the dataset provided by~\cite{tang2021ganda}. The experimental results suggest that DAMM-Diffusion can generate the distribution of NPs with higher accuracy than the comparing methods. We summarize the main contributions of this study as follows: 

\begin{itemize}
    \item We propose a novel diffusion-based multi-modal generative model (i.e., DAMM-Diffusion) consisting of both uni-modal and multi-modal branches for NPs distribution prediction. 
    \item To our best knowledge, we first consider the divergence among different modalities in the multi-modal image generation task. The proposed Divergence-Aware Multi-Modal Predictor (DAMMP) can adaptively generate the prediction results from uni-modal or multi-modal branches.
    \item We introduce the Multi-Modal Fusion Module (MMFM) and Uncertainty-Aware Fusion Module (UAFM) in the multi-modal branch that can better fuse the information of different modalities.
    \item Experimental results for both internal and external dataset validations as well as on the brain image synthesis task suggest the advantages of our  model in comparison with other methods.
\end{itemize}

\begin{figure*}[htbp]
  \centering
  \includegraphics[width=0.9\linewidth]{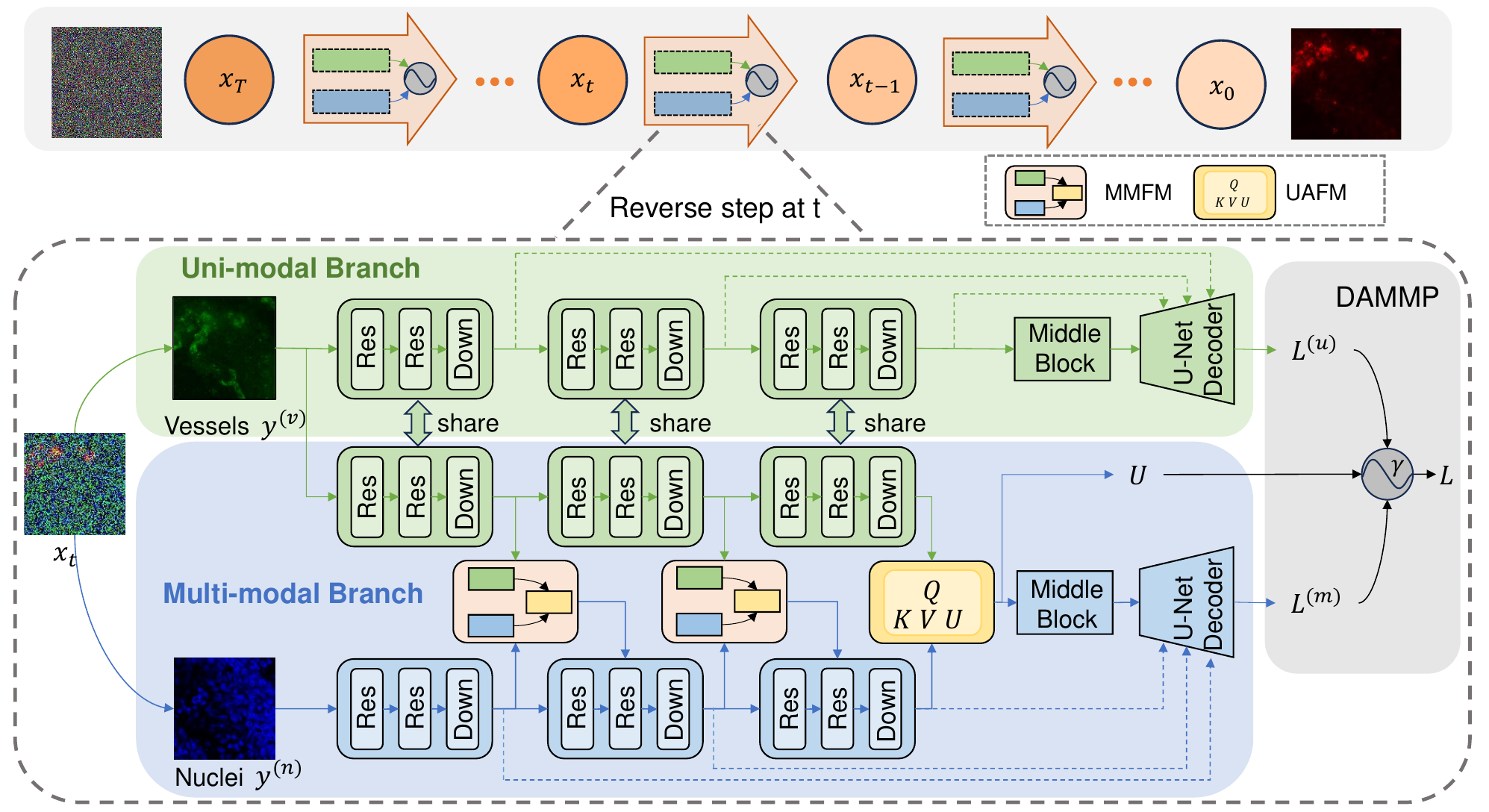}
  \caption{Overview of DAMM-Diffusion. At each step of the reverse process, both the uni-modal branch and multi-modal branch perform the reverse step in a unified network, and output the predictions respectively. The uni-modal branch consists of a U-Net architecture with the encoder, middle U-Net bottleneck and a decoder. The multi-modal branch additionally incorporates two modules i.e., MMFM and UAFM aiming at fusing the multi-modal data. Finally, DAMMP decides whether to apply multi-modal predictions or only using the uni-modal generation results for NPs distribution prediction.}
  \label{fig:fig2}
  \vspace{-1.0em}
\end{figure*}

\section{Related Work}
\paragraph{Diffusion Models.} Diffusion Models~\cite{ho2020denoising,song2020score} have emerged as a new type of generative models, achieving impressive results across different applications like computer vision~\cite{dhariwal2021diffusion,rombach2022high,chen2023diffusiondet}, nature language processing~\cite{gong2022diffuseq,li2022diffusion}, video synthesis~\cite{ge2023preserve,ruan2023mm}, etc. It is capable of generating high-quality and detailed images by just using a single mean squared error (MSE) loss to optimize the lower variational bound on the likelihood function. Recently, the diffusion models have gained significant prominence in image-to-image translation (I2I) tasks~\cite{choi2021ilvr,saharia2022image,rombach2022high,li2023bbdm}. For example, LDM~\cite{rombach2022high} designs a general-purpose conditioning mechanism based on cross-attention by applying the diffusion model on the latent space to enhance I2I task efficiency. BBDM~\cite{li2023bbdm} learns the translation between two domains directly through the bidirectional diffusion process, which is called a stochastic Brownian Bridge process.
In this paper, we view the NPs distribution prediction as the I2I task, and apply the diffusion model to solve it. Specifically, our DAMM-Diffusion integrates nuclei (stained by DAPI) and vessels (stained by CD31) images for the prediction of NPs distribution. \\
\textbf{Multi-Modal Image Generation.} Multi-modal image generation aims to generate images from the multi-modal inputs, including images with multi-modality, image-text pairs, etc. Generally, the existing multi-modal image generation models can be divided into two categories, i.e., the generative adversarial network (GAN) based models and the diffusion model based approaches. 
For the GAN based models, MM-GAN~\cite{sharma2019missing} is capable of synthesizing missing MRI sequences as a variant of pix2pix model~\cite{isola2017image} by combining information from all available sequences. DiamondGAN~\cite{li2019diamondgan} designs a multi-modal cycle-consistency loss function to better learn the multiple-to-one cross-modality mapping. ResViT~\cite{dalmaz2022resvit} introduces a transformer-based generator to translate between multi-modal imaging data. 
Unlike GANs, diffusion models employ a probabilistic approach to model the underlying distribution of the training data and have also been applied to multi-modal image generation tasks.
For instance, DiffDP~\cite{feng2023diffdp} introduces a structure encoder to extract the anatomical information and apply it to guide the noise predictor in the diffusion model. CoLa-Diff~\cite{jiang2023cola} is a conditional latent diffusion model can effectively balance multiple conditions. In addition to the methods using multi-modal image as inputs, recent studies also incorporate image-text pairs and can  achieve impressive results~\cite{huang2022multimodal,kumari2023multi,voynov2023sketch}. However, to the best of our knowledge, all these multi-modal generative models do not take the divergence among the input multi-modal data into consideration, which may lead to sub-optimal performance.

\section{Method}
\subsection{Problem Formulation}
We generate NPs using tumor vessels and cell nuclei since they are highly correlated with NPs distribution~\cite{tang2021ganda}. Mathematically, 
given the input paired data ($y^{(v)}, y^{(n)}\in R^{H\times W}$) from the image set of tumor vessels $\mathcal{V}$ ($y^{(v)}\in\mathcal{V}$) and cell nuclei $\mathcal{N}$ ($y^{(n)}\in\mathcal{N}$), our goal is to generate NPs  i.e., $x\in R^{H\times W}$ on the condition of $y^{(v)}$ and $y^{(n)}$.

\subsection{Divergence-Aware Multi-Modal Diffusion Models for NPs Distribution Prediction}
Different from the previous work \cite{xu2023high} which directly predicts the target based on the input multi-modal images, our DAMM-Diffusion Model considers both the uni-modal and multi-modal branches within one diffusion process, and determines whether to integrate vessels and nuclei information for NPs distribution prediction since the divergence between them may negatively impact the final output. Following the prior work~\cite{rombach2022high}, we take the diffusion process in the latent space of VAE/VQGAN for efficiency. For simplicity of notations, we still use $y^{(v)}$, $y^{(n)}$, $x$ to denote the latent features of vessels, nuclei and generated NPs distribution, respectively.

In the latent space, the $T$-step forward process progressively adds noise to the original NPs distribution $x_0\sim q(x_0)$, which can be described as follows:
\begin{equation}
q(x_1,...,x_T|x_0)=\prod_{t=1}^{T}q(x_t|x_{t-1}),
\label{eq:dammp_forward}
\end{equation}
where $q(x_t|x_{t-1})=\mathcal{N}(x_t;\sqrt{1-\beta_{t}}x_{t-1}, \beta_{t}\mathbf{I})$ is a normal distribution whose mean value is $\sqrt{1-\beta_{t}}x_{t-1}$ and the deviation is $\beta_t\mathbf{I}$. Here, $\beta_t$ is the variance schedule across diffusion steps. The latent variable $x_{T}\sim\mathcal{N}(0,\mathbf{I})$  when $T{\rightarrow}\infty$.

In the reverse process, the uni-modal and multi-modal branches perform their reverse steps, conditioned on uni-modal and multi-modal data, respectively. 
Specifically, the uni-modal branch is injected with the tumor vessel image  $y^{(v)}$ and can be formulated as: 
\begin{equation}
    \resizebox{.91\linewidth}{!}{$
            \displaystyle
        p^{(u)}_{\theta}(x_{t-1}|x_{t},y^{(v)})= \mathcal{N}(x_{t-1};\mu^{(u)}_{\theta}(x_t,y^{(v)}_t,t), \Sigma^{(u)}_{\theta}(x_t,y^{(v)}_t,t))
        $}.
\end{equation}
Similarly, the multi-modal branch is associated with both $y^{(v)}$ and the nuclei image $y^{(n)}$ defined as:
\begin{equation}
    \resizebox{.91\linewidth}{!}{$
            \displaystyle
        p^{(m)}_{\theta}(x_{t-1}|x_{t},y^{(v)},y^{(n)})= 
        \mathcal{N}(x_{t-1};\mu^{(m)}_{\theta}(x_t,y^{(v)}_t,y^{(n)},t), \Sigma^{(m)}_{\theta}(x_t,y^{(v)}_t,y^{(n)},t))
        $}.
\end{equation}
Instead of separately modeling the $\theta_u$ and $\theta_{m}$ for $p^{(u)}$ and $p^{(m)}$, we perform both uni-modal and multi-modal branches in a unified network $\theta$ (i.e., UMM-Network) to obtain their outputs \emph{i.e., } $\epsilon_{\theta}^{(u)}$ and $\epsilon_{\theta}^{(m)}$ with the following objectives:
\begin{equation}
\mathcal{L}^{(u)} = \mathbb{E}_{x_0,y^{(v)},t,\epsilon}||\epsilon-\epsilon^{(u)}_{\theta}(x_t,y^{(v)},t)||^2_2
\label{eq:loss_u}
\end{equation}
\begin{equation}
\mathcal{L}^{(m)} = \mathbb{E}_{x_0,y^{(v)},y^{(n)},t,\epsilon}||\epsilon-\epsilon^{(m)}_{\theta}(x_t,y^{(v)},y^{(n)},t)||^2_2
\label{eq:loss_m}
\end{equation}
In the last stage of each reverse step, the Divergence-Aware Multi-Modal Predictor (DAMMP) determines the final output $\epsilon_{o}$ from $\epsilon_{\theta}^{(u)}$ and $\epsilon_{\theta}^{(m)}$ based on the divergence value described in \cref{sec:method_DAMMP}.


\subsection{The Unified Network Combining Both Uni-Modal and Multi-Modal Branches.}
In our UMM-Network, the uni-modal branch takes vessel images as input and is based on U-Net backbone that is proven to be effective in diffusion models~\cite{ho2020denoising,nichol2021improved}. 
As depicted in \cref{fig:fig2}, the multi-modal branch also employs the similar U-Net backbone but additionally incorporates two fusion modules i.e., the multi-modal fusion module (MMFM) and uncertainty-aware fusion module (UAFM). 
In both branches, the time step $t$ is shared, and the condition images $y^{(v)}$ and $y^{(n)}$ are separately concatenated with the $t$-step noisy image $x_t$, forming a new feature pair $(v, n)$ before feeding into their corresponding encoders. 

\begin{figure}[htbp]
  \centering
  \includegraphics[width=\linewidth]{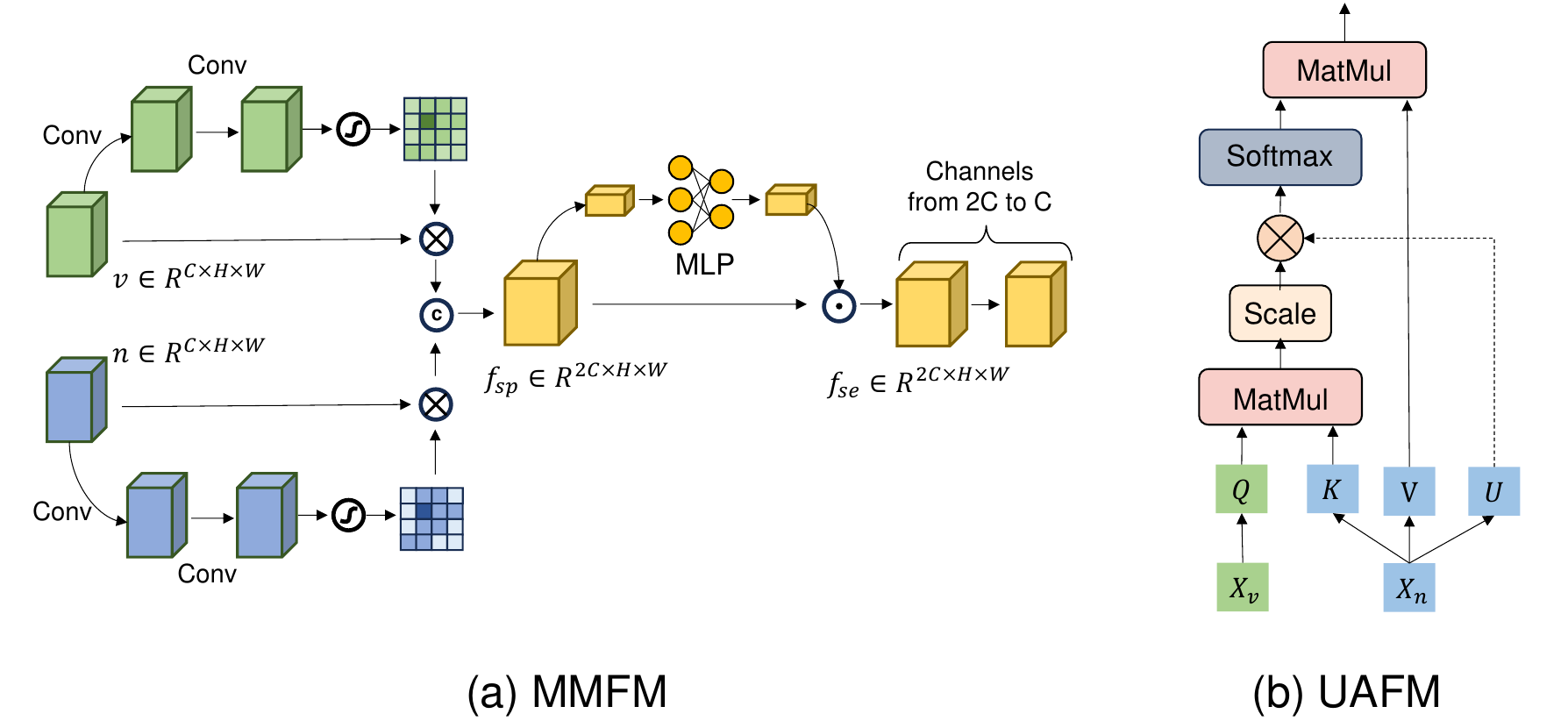}
  \caption{Illustration of the proposed Multi-Modal Fusion Module (MMFM) and Uncertainty-Aware Fusion Module (UAFM).}
  \label{fig:fig3}
\end{figure}


\subsubsection{Multi-Modal Fusion Module}
To effectively fuse the features derived from two modalities, we introduce the Multi-Modal Fusion Module (MMFM). As illustrated in \cref{fig:fig3} (a), MMFM firstly extracts the useful spatial information from individual modality, and then reweights the channels after concatenating them together. Finally, the $1\times 1$ convolution is applied to compress the channels. More specifically, taking the tumor vessels feature $v\in R^{C\times H \times W}$ as an example, the spatial attention weights can be computed as: 
\begin{equation}
w_{v} = \sigma(\mathcal{B}(Conv(\delta(\mathcal{B}(Conv(v)))))),
\label{eq:mmfm_1}
\end{equation}
where $\sigma$ is the Sigmoid function, $\mathcal{B}$ denotes the Batch Normalization (BN), and $\delta$ denotes the Rectified Linear Unit (ReLU). The spatial attention output can be represented as $v_{sp} = w_v\otimes v$, where $\otimes$ refers to element-wise multiplication. We can similarly get $n_{sp}$, which is the output of spatial attention from the modality of cell nuclei. Next, for the purpose of combining the features from different modalities, we combine $v_{sp}\in R^{C\times H\times W}$ and $n_{sp}\in R^{C\times H\times W}$ as $f_{sp}=[v_{sp}, n_{sp}] \in R^{2C\times H\times W}$. Then, the channel attention weights of the concatenated feature can be computed by:
\begin{equation}
w_{se}=\sigma(\mathcal{B}(W_{2}\delta(\mathcal{B}(W_{1}Avgpool(f_{sp}))))),
\label{eq:mmfm_2}
\end{equation}
where $W_{1}\in R^{\frac{2C}{r} \times 2C}$, $W_{2}\in R^{2C\times \frac{2C}{r}}$, and $r$ is a reduction ratio which help limit model complexity. Based on \cref{eq:mmfm_2}, the output of the MMFM module can be denoted as $f_{se}=w_{se}\odot f_{sp}$ where $\odot$ refers to channel-wise multiplication. Finally, a $1\times 1$ convolution help compress the channels of $f_{se}$ from $2C$ to $C$.

\subsubsection{Uncertainty-Aware Fusion Module}
\label{sec:UAFM}
As shown in \cref{fig:fig2}, besides the MMFM modules applied in the multi-branch of our model, we develop an Uncertainty-Aware Module (UAFM) that can calculate the uncertainty map for multi-modal fusion during cross-attention computation. Specifically, given the input feature maps $X_v\in R^{H \times W\times C}$ from the modality of tumor vessels and $X_n\in R^{H \times W\times C}$ from the modality of cell nuclei, the traditional cross-attention can be formulated via: 
\begin{equation}
CA(Q,K,V)=softmax(\frac{QK^T}{(\sqrt{d})})V,
\end{equation}
where $Q=W_{q}X_{v}$, $K=W_{k}X_{n}$ and $V=W_{v}X_{n}$. $W_{q},W_{k},W_{v}\in R^{C\times d}$. Inspired by \cite{ying2022uctnet}, the original cross-attention can be interpreted as a bipartite graph where $V$ (from the modality of cell nuclei) represents the node features and $QK^{T}$ is the attention weights in the form of adjacency matrix by measuring the similarity between $Q$ (from the modality of tumor vessels) and $K$ (from the modality of cell nuclei) that can instruct each node updating its information from the neighborhood nodes. However, attention weights are not always reliable since the existence of uncertain nodes. Our UAFM considers such uncertainty and learns to measure it. Specifically, we define a learnable uncertainty map $U\in R^{H\times W\times 1}$ to represent the uncertainty of each node, and the uncertainty value on the $k$-$th$ node affects its edge set $\{\varepsilon_{i\rightarrow j}\}_{i=k}$, such that the information flow from the node $k$ propagates with an uncertainty value. Finally, the uncertainty-aware cross-attention can be formulated as: 
\begin{equation}
UACA(Q,K,V,U)=softmax(\frac{QK^T \cdot (1-U)}{(\sqrt{d})})V,
\label{eq:uaca}
\end{equation}
and the uncertainty map $U$ can be computed as:
\begin{equation}
    U=W_{n}X_n,
\label{eq:U}
\end{equation}
where $W_n\in R^{C\times 1}$ can be learned from the network. The uncertainty map $U$ assesses the uncertainty value of each node, and the node with larger uncertainty value will have lower impact on the edges set $\{\varepsilon_{i\rightarrow j}\}_{i=k}$, and vice versa.

\subsection{Divergence-Aware Multi-Modal Predictor}
\label{sec:method_DAMMP}
The existing study \cite{yang2022domfn} indicates that the divergence among multi-modal data may lead to unreliable results when integrating them together. Accordingly, it is necessary to decide whether to rely on the multi-modal predictions or leverage uni-modal results. On the other hand, learning a multi-modal network that incorporates both uni-modal and multi-modal branches requires a unified training method, rather than tackling the uni-modal or multi-modal branch separately. To address the above two issues, we propose a Divergence-Aware Multi-Modal Predictor (DAMMP) to assess the consistency within the multi-modal data. Specifically, we calculate the mean value of $U$ (\cref{eq:U}), denoted as $d$, to represent the divergence for the current multi-modal data. Intuitively, a high value of $d$ indicates the lack of consistency in the multi-modal fusion process since it refers to low confidence in the attention weights of cross-attention. In such case, the divergence of multi-modal data will affect the fusion process and the results come from uni-modal branch is more reliable. In other words, we just optimize the uni-modal loss if $d$ is larger to the threshold $\gamma$. By contrast, we encourage multi-modal fusion when the divergence $d$ is lower than a threshold $\gamma$, indicating high confidence in cross-attention and the potential for reliable and consistent multi-modal fusion. In this case, we optimize the loss functions for both uni-modal (\cref{eq:loss_u}) and multi-modal branches (\cref{eq:loss_m}). In summary, the formulation of the overall loss can be represented as: 
\begin{equation}
\mathcal{L}=
\begin{cases}
\mathcal{L}^{(u)}+\mathcal{L}^{(m)}& d\le \gamma \\
\mathcal{L}^{(u)}& d>\gamma
\end{cases}
\label{eq:dammp1}
\end{equation}
Although \cref{eq:dammp1} provides a flexible way for the combination of multi-modal predictors, the choice of $\gamma$ may significantly influence the final prediction results. To enhance the model's adaptability on $\gamma$, we dynamically feed the results of multi-modal fusion back to the divergence $d$. For this purpose, we design a divergence feedback loss (DFL). Specifically, we firstly compute the $L_1$ distance between the outputs of the two branches and the groundtruth $\epsilon$, i.e., $L_{1}^{(u)}=||\epsilon-\epsilon^{(u)}_{\theta}||_{1}$ and $L_{1}^{(m)}=||\epsilon-\epsilon^{(m)}_{\theta}||_{1}$. Then, we design a criterion to represent the effectiveness of multi-modal fusion below: 
\begin{equation}
e=
\begin{cases}
1& L_{1}^{(m)}\le L_{1}^{(u)} \\
0& L_{1}^{(m)}> L_{1}^{(u)}
\end{cases}
\end{equation}
Here $e$ is an indicator referring to whether the multi-modal branch performs better than the uni-modal branch. In detail, DFL can be defined as the binary cross entropy loss with a regularization term:
\begin{equation}
\mathcal{L}_{dfl}=e\log (1-d)+(1-e)\log d + \alpha||d-\gamma||_2^2,
\label{eq:dflloss}
\end{equation}
where $\alpha$ is a tuning parameter and the regularization term is applied to ensure a stable divergence value $d$.  
Finally, the total loss of our DAMM-Diffusion can be formulated as: 
\begin{equation}
\mathcal{L}^{total} = \mathcal{L} + \lambda \mathcal{L}_{dfl},
\label{eq:totalloss}
\end{equation}
where $\lambda$ controls the relative importance of DFL.

In the prediction phase, only one output is employed in the reverse step. The final output is the prediction from multi-modal branch when $d$ is lower than $\gamma$. In contrast, when $d$ is high, which means that the confidence of multi-modal fusion is low, the output from uni-modal branch is selected. In summary, the final output $\epsilon_{o}$ can be represented as: 
\begin{equation}
\epsilon_{o}=
\begin{cases}
\epsilon^{(m)}_{\theta}& d\le \gamma \\
\epsilon^{(u)}_{\theta}& d> \gamma
\end{cases}
\end{equation}

\section{Experiments}
\label{sec:exp}
\subsection{Experimental Setup}

\paragraph{Datasets.} To evaluate the  performance of our DAMM-Diffusion, we conduct the experiments on the dataset introduced in ~\cite{tang2021ganda} that predicts the NPs distribution pixels-to-pixels given the TME components of tumor vessels and cell nuclei. Here, the NPs distribution is visualized by injecting 20-nm PEGylated CdSe/ZnS quantum dots (QDs) into T1 mice models of breast cancer. The tumor blood vessels are labeled with a primary CD31 antibody, while the cell nuclei are identified by DAPI staining. 
Following the prior work~\cite{tang2021ganda, xu2023high}, we conduct cross-validation for  internal validation and further test our model on the external validation.
For the internal validation, a total of 25,097 patches are collected, and a six-fold cross-validation is performed, with each fold corresponding to an individual T1 mice model. The external validation dataset is consisted of 3,800 patches derived from a different B16 tumor model in order to further validate the generalization ability of our model. For the purpose of further validating the effectiveness of our method, another brain image synthesis dataset i.e., BRATS dataset~\cite{menze2014multimodal} is adopted. More details for the BRATS dataset is presented in the \emph{Supplementary Material}.

\vspace{-3mm}
\paragraph{Implementation Details.} 
We adopt a linear noise schedule in our DAMM-Diffusion. The number of time steps is set to 1000 during the training stage, and we use 150 sampling steps during the inference stage to balance sample quality and efficiency. The hyper-parameters in DAMMP are set as follows: $\alpha$ to 0.1, $\lambda$ to 1e-4 and $\gamma$ to 0.5. More discussions for the different values of these parameters can be found in the \emph{Supplementary Material}. All of our experiments are implemented in PyTorch through the NVIDIA RTX 3090 GPU. \\

\begin{table}[htbp]
    \vspace{-2mm}
    \centering
    \caption{Performance comparisons with state-of-the-art methods on internal validation set. The symbol $^{*}$ indicates significant improvement ($p<$ 0.05).}
    \resizebox{\linewidth}{!}{
    \begin{tabular}{cc|cc}
        \hline
        \multicolumn{2}{c|}{Methods} & SSIM \% & PSNR\\
        \hline
        \multirow{4}{*}{\parbox{1.8cm}{Uni-modal}}
         &Cyclegan & 84.07$\pm$2.67 $^{*}$ & 36.96$\pm$2.34 $^{*}$ \\
         &Pix2pix & 87.81$\pm$2.15 $^{*}$ & 38.97$\pm$2.70 $^{*}$ \\
         &LDM & 92.97$\pm$0.65 $^{*}$ & 43.72$\pm$0.62 $^{*}$\\
         &BBDM & 93.01$\pm$0.81 $^{*}$ & 43.96$\pm$0.75 $^{*}$ \\ 
         \hline
         \multirow{9}{*}{\parbox{1.8cm}{Multi-modal}}
         &DiamondGAN & 88.14$\pm$2.66 $^{*}$  & 41.84$\pm$2.61 $^{*}$ \\
         &GANDA & 89.53$\pm$2.37 $^{*}$ & 42.83$\pm$2.02 $^{*}$ \\
         &HiNet & 90.83$\pm$1.82 $^{*}$ & 42.78$\pm$1.60 $^{*}$  \\
         &ResViT & 91.88$\pm$1.26 $^{*}$  & 44.14$\pm$1.27 $^{*}$ \\
         &HRPN &  93.32$\pm$0.93 $^{*}$  & 44.37$\pm$1.02 $^{*}$  \\
         &Collaborative & 92.96$\pm$1.08 $^{*}$  & 43.52$\pm$0.82 $^{*}$  \\
         &DiffDP & 94.21$\pm$0.78 $^{*}$ & 44.86$\pm$0.66 $^{*}$  \\
         &CoLa-Diff & 94.36$\pm$0.74 $^{*}$   & 45.02$\pm$0.69 $^{*}$  \\
         &Ours & \textbf{96.54$\pm$0.62} & \textbf{47.93$\pm$0.67}\\
        \hline
    \end{tabular}}\\
    \vspace{-2mm}
    \label{tab:table1}
\end{table}

\begin{table}[htbp]
    \centering
    \caption{Performance comparisons with state-of-the-art methods on external validation set. The symbol $^{*}$ indicates significant improvement ($p<$ 0.05).}
    \resizebox{\linewidth}{!}{
    \begin{tabular}{cc|cc}
        \hline
        \multicolumn{2}{c|}{Methods} & SSIM \% & PSNR\\
        \hline
        \multirow{4}{*}{\parbox{1.8cm}{Uni-modal}}
         &Cyclegan & 79.74$\pm$3.21  $^{*}$ & 35.84$\pm$3.34  $^{*}$\\
         &Pix2pix & 81.75$\pm$2.55  $^{*}$ & 36.84$\pm$2.87  $^{*}$ \\
         &LDM & 83.22$\pm$1.22 $^{*}$ & 37.52$\pm$1.20 $^{*}$ \\
         &BBDM & 83.86$\pm$1.21  $^{*}$ & 38.23$\pm$1.04  $^{*}$ \\
         \hline
         \multirow{9}{*}{\parbox{1.8cm}{Multi-modal}}
         &DiamondGAN & 83.25$\pm$2.94  $^{*}$ & 38.63$\pm$2.35 $^{*}$ \\ 
         &GANDA & 84.24$\pm$2.59 $^{*}$ & 38.62$\pm$2.06 $^{*}$ \\
         &HiNet & 85.95$\pm$2.39  $^{*}$  & 39.61$\pm$1.86 $^{*}$  \\
         &ResViT & 85.95$\pm$1.73  $^{*}$ & 39.34$\pm$1.28  $^{*}$\\
         &HRPN & 86.29$\pm$1.33 $^{*}$  & 39.81$\pm$1.30  $^{*}$\\
         &Collaborative & 85.85$\pm$1.67  $^{*}$ & 39.72$\pm$1.65  $^{*}$ \\
         &DiffDP & 86.20$\pm$1.25  $^{*}$ & 40.12$\pm$1.16 $^{*}$ \\
         &CoLa-Diff & 86.72$\pm$1.14 $^{*}$  & 40.28$\pm$1.37 $^{*}$  \\
         &Ours & \textbf{88.79$\pm$1.07 } & \textbf{41.95$\pm$0.96 } \\
        \hline
    \end{tabular}}\\
    \label{tab:table3}
    \vspace{-3mm}
\end{table}
\vspace{-5mm}
\paragraph{Comparison Methods and Evaluation Metrics:} To evaluate the quality of NPs distribution generated by DAMM-Diffusion, we compare it with both uni-modal image translation methods such as Pix2pix~\cite{isola2017image}, Cyclegan~\cite{zhu2017unpaired}, LDM~\cite{rombach2022high}, BBDM~\cite{li2023bbdm} and multi-modal image generation methods including GANDA~\cite{tang2021ganda}, HRPN~\cite{xu2023high}, Hi-Net~\cite{zhou2020hi}, ResViT~\cite{dalmaz2022resvit}, DiffDP~\cite{feng2023diffdp}, Cola-Diff~\cite{jiang2023cola} and Collaborative Diffusion~\cite{huang2023collaborative}. We employ the metrics of Peak Signal to Noise Ratio (PSNR) and Structural Similarity (SSIM) to evaluate the generation performance of different methods.

\subsection{Comparison with State-of-the-art Methods} \label{sec:exp_comparison}
\paragraph{Evaluation on the Internal Validation Set.} 
We firstly compare our DAMM-Diffusion with SOTA uni-modal and multi-modal generative models (shown in \cref{tab:table1}) on the internal validation set. For the uni-modal methods, we only report the best individual prediction results among the nuclei and vessel channels. Additional results for each individual modality of uni-modal methods are provided in the \emph{Supplementary Material}. As shown in \cref{tab:table1}, the proposed DAMM-Diffusion significantly outperforms the comparing methods on each individual modality data, these results clearly demonstrate the advantage of combining the nuclei and vessels for NPs distribution prediction. Next, no matter adopting the uni-modal or multi-modal models for NPs generation, the studies based on the diffusion models (i.e., BBDM and CoLa-Diff) can consistently yield better results than their competitors, demonstrating the rationality of our methods built on the diffusion model. In addition, although the multi-modal image translation models are superior to the uni-modal studies in general, we can find some exceptions. For instance, the SOTA uni-modal methods such as LDM and BBDM are superior to several multi-modal generative methods like DiamondGAN and GANDA. One possible reason is that these multi-modal algorithms overlook the divergence among different modalities, leading to the inconsistent fusion results. Finally, our method outperforms the SOTA uni-modal methods and multi-modal methods in terms of PSNR and SSIM, verifying its ability to effectively fuse both uni-modal branch and multi-modal branch predictions, showcasing its reliability and superiority.  \\
\textbf{Evaluation on the External Validation Set.} To further explore the generalization and effectiveness of our DAMM-Diffusion, we compare it with SOTA methods for the external validation, and the results are shown in \cref{tab:table3}. As observed from \cref{tab:table3}, the approaches combing both vessel and nuclei channels can achieve better prediction results than the uni-modal studies. These results clearly demonstrate the advantage of the integrative analysis of multi-modal TME components for NPs generation. Moreover, our DAMM-Diffusion outperforms other SOTA multi-modal image translation methods, which further affirms the effectiveness of our proposed divergence-aware diffusion model with strong generalization ability.

\subsection{Qualitative Results}
\begin{figure}[htbp]
  \centering
  \includegraphics[width=0.95\linewidth]{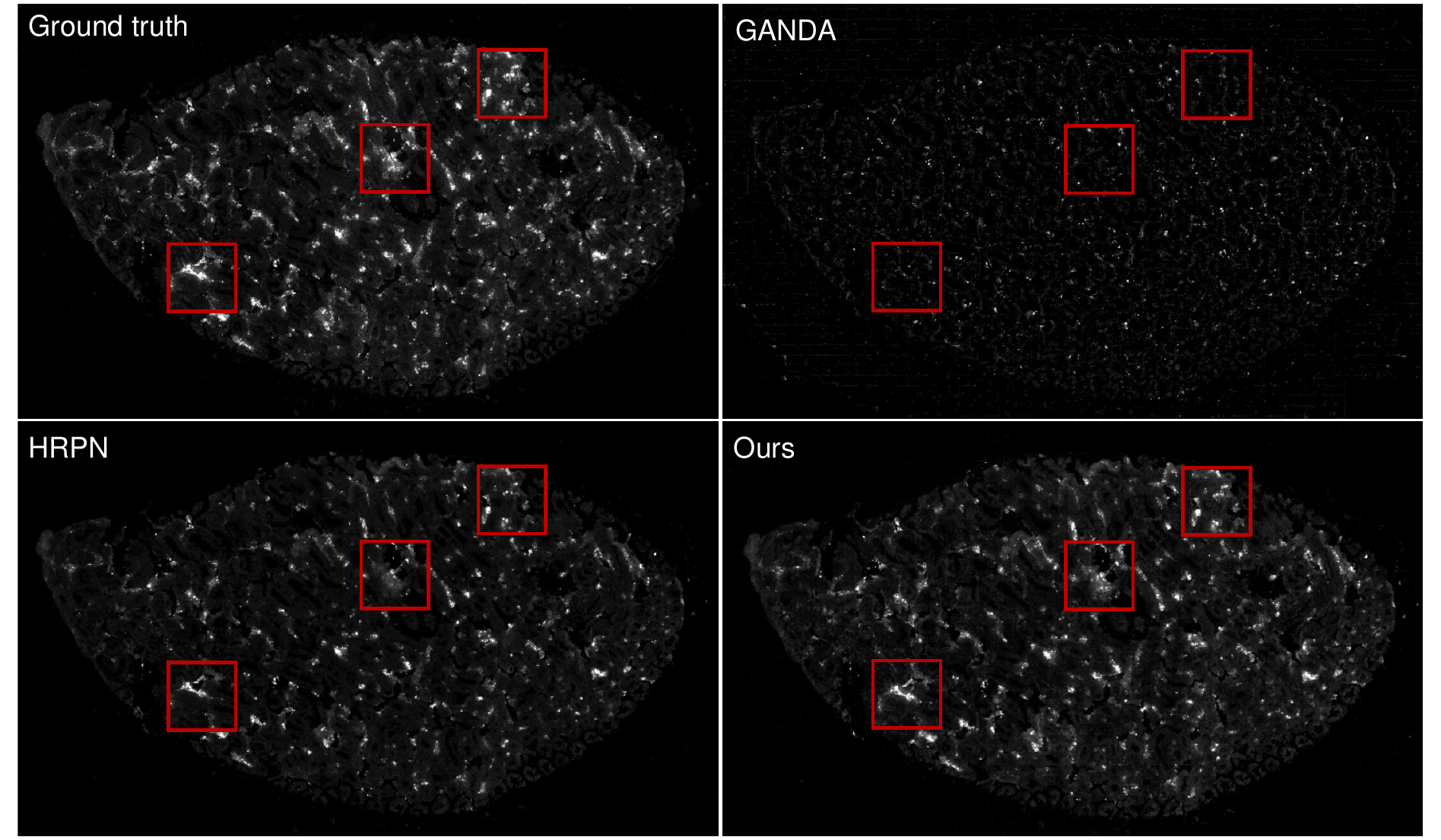}
  \caption{Qualitative comparison between the proposed method and the previous methods for NPs distribution at whole-slide level.}
  \label{fig:fig_ws}
  \vspace{-4mm}

\end{figure}

\begin{figure}[htbp]
  \centering
  \includegraphics[width=\linewidth]{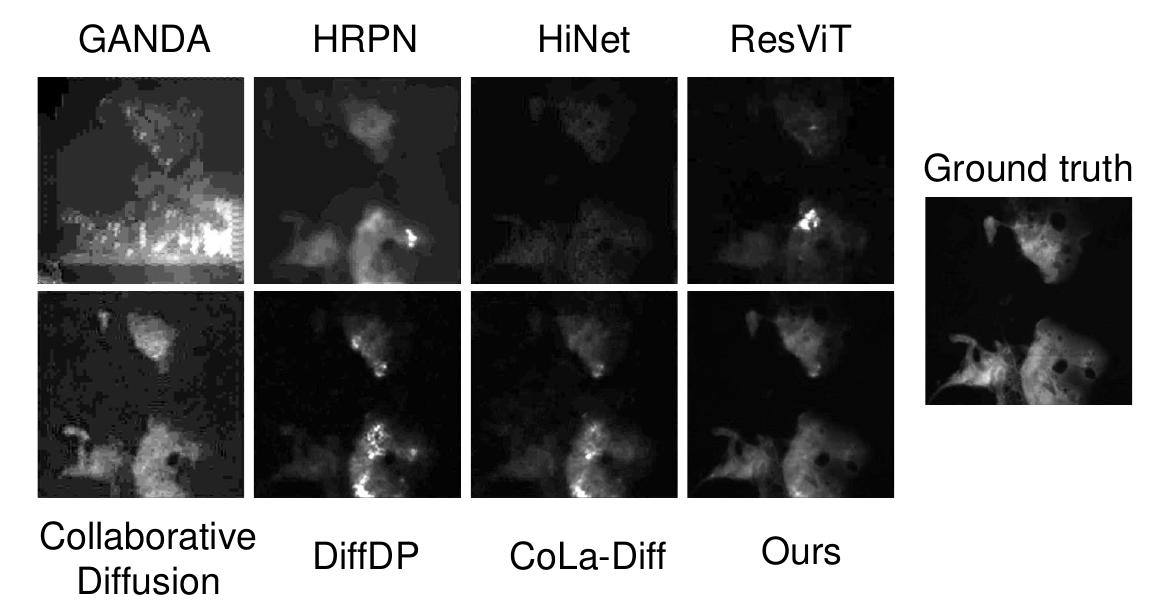}
  \caption{Qualitative comparison between the proposed method and other state-of-the-art methods at patch-level.}
  \label{fig:fig_patch}
  \vspace{-4mm}
\end{figure}

In this section, we compare the whole-slide level and patch-level visualization results of NPs with its competitors in Fig.~\ref{fig:fig_ws} and Fig.~\ref{fig:fig_patch}, respectively. As shown in  Fig.~\ref{fig:fig_ws}, we note that the NPs distribution predicted by our DAMM-Diffusion is closer to the ground truth than the comparing methods at WSI-level. On the other hand, as indicated in Fig.~\ref{fig:fig_patch}, we observe that our DAMM-Diffusion is also effective in preserving smooth boundaries and maintaining the NPs distribution at patch-level. Additional results are provided in the \emph{Supplementary Material} for easier viewing.

\newcommand{\TtwoFlairTone}{T\textsubscript{2},~FLAIR~$\rightarrow$~T\textsubscript{1}}
\newcommand{\ToneFlairTtwo}{T\textsubscript{1},~FLAIR~$\rightarrow$~T\textsubscript{2}}
\newcommand{\ToneTtwoFlair}{T\textsubscript{1},~T\textsubscript{2}~$\rightarrow$~FLAIR}

\renewcommand{\tabcolsep}{2pt}
\renewcommand{\arraystretch}{1.20}
\begin{table}[htbp]
\centering
\resizebox{1.02\columnwidth}{!}{%
\begin{tabular}{ccccccc}
\hline
\multirow{2}{*}{Methods}          & \multicolumn{2}{c}{\ToneTtwoFlair}  & \multicolumn{2}{c}{\ToneFlairTtwo} & \multicolumn{2}{c}{\TtwoFlairTone}  \\ \cline{2-7} 
& PSNR& SSIM \%& PSNR& SSIM \%& PSNR& SSIM \%  \\ \hline
\multirow{2}{*}{DiamondGAN}& 22.18 & 81.57
 & 23.82& 87.32& 23.92 &  87.40\\
&$\pm$3.24 &$\pm$5.67 & $\pm$3.25 &  $\pm$5.13 & $\pm$2.69  &   $\pm$3.80\\ \hline
\multirow{2}{*}{GANDA}
&22.81 &   82.20& 23.96&  86.80&   23.94&  88.66 \\
&$\pm$3.05&  $\pm$5.41&  $\pm$3.17&   
$\pm$4.88&   $\pm$2.77&     $\pm$4.44\\ \hline
\multirow{2}{*}{HiNet}&     
21.58& 	84.94&	23.22& 	88.02&	23.48& 	89.71 \\
&$\pm$3.87&  $\pm$5.23&  $\pm$3.31&   $\pm$3.26&   $\pm$2.31&     $\pm$3.12\\ \hline
\multirow{2}{*}{ResViT} &      
23.64& 	86.01&	24.30& 	90.30&	23.18& 	88.14   \\
&$\pm$3.22&  $\pm$4.48&  $\pm$3.62&   $\pm$3.03&   $\pm$2.97&     $\pm$4.95    \\ \hline
\multirow{2}{*}{HRPN}&        
23.26& 	86.27&	23.43& 	89.96&	23.83& 	89.52\\
&$\pm$2.53&  $\pm$4.92&  $\pm$3.10&   
$\pm$3.26&   $\pm$2.59&  $\pm$2.06         \\ \hline
\multirow{2}{*}{Collaborative}     &     
22.90& 	86.20&	22.96& 	89.93&	23.59& 	89.34\\
&$\pm$3.38 &  $\pm$5.08 &  $\pm$3.66
&$\pm$3.50  &   $\pm$2.56&     $\pm$3.47\\ \hline
\multirow{2}{*}{DiffDP} &      
23.07& 	87.40&	23.88& 	91.29&	24.43& 	91.24\\
&$\pm$3.37&  $\pm$6.01&  $\pm$3.66&   
$\pm$4.00&   $\pm$2.55&     $\pm$3.42 \\ \hline
\multirow{2}{*}{CoLa-Diff} &      
23.37& 	88.04&	24.19& 	91.62&	24.69& 	91.52\\
&$\pm$3.35&  $\pm$5.65 &  $\pm$3.62         
&$\pm$3.81 &   $\pm$2.49&     $\pm$3.29         \\ \hline    
\multirow{2}{*}{Ours} &      
\textbf{24.23}& \textbf{89.20}&	\textbf{25.25}& 	\textbf{92.86}&	\textbf{25.61}& 	\textbf{92.68}\\
&\textbf{$\pm$3.16}&\textbf{$\pm$5.65}&\textbf{$\pm$3.48} 
&\textbf{$\pm$3.60}&\textbf{$\pm$2.49}&\textbf{ $\pm$3.22}  \\ \hline  
\end{tabular}}
\caption{
Performance of task-specific synthesis models in many-to-one tasks (\ToneTtwoFlair, \ToneFlairTtwo, and \TtwoFlairTone) across test subjects in the BRATS dataset. Boldface indicates the top-performing model for each task.}
\label{tab:brats_many}
\vspace{-3mm}
\end{table}

\vspace{-2mm}
\subsection{Applicability to Brain Image Synthesis Task}
We further validate the applicability of our method to brain image synthesis task. For the multi-modal brain images (i.e., T\textsubscript{1},  T\textsubscript{2} and FLAIR) in the BRATS dataset, we iteratively generate the brain image of specific modality given the image with other modalities. The results are shown in \cref{tab:brats_many}. For clarity, we only report the best performance of DAMM-Diffusion on uni-modal branch (more results can be found in the \emph{Supplementary Material}). As shown in \cref{tab:brats_many}, DAMM-Diffusion still significantly outperforms ($p<0.05$) other comparing methods by the measurements of PSNR and SSIM. These results clearly demonstrate the applicability of our model to other multi-modal image generation tasks. More visualization results for the brain image synthesis tasks are provided in the \emph{Supplementary Material}.

\subsection{Ablation Study}
\paragraph{Effects of Choosing Different Types of Images in the Uni-Modal Branch.} In our DAMM-Diffusion, the inputs of the uni-modal branch are vessel images since both the previous studies \cite{sulheim2018multi,tang2021ganda} and the experimental results shown in \cref{tab:table1} indicate that they are more relevant to the distribution of NPs. In order to illustrate the advantage of choosing the vessel images in the uni-modal branch, we further compare DAMM-Diffusion with its variant that applies nuclei images in the uni-modal branch. The experimental results shown in \cref{tab:ablation_modality} clearly demonstrate that the prediction of NPs distribution can be surely improved if the inputs of the uni-modal branch are vessel images.  

\begin{table}[htbp] 
    \centering
    \caption{Effects of choosing different types of images in the uni-modal branch.}
    \vspace{-2mm}
    \resizebox{\linewidth}{!}{
       \begin{tabular}{c|cc| c c} 
       \hline
       {\multirow{2}{*}{Input images}} &\multicolumn{2}{c|}{Internal validation}&\multicolumn{2}{c}{External validation} \\
        & SSIM \% & PSNR & SSIM \% & PSNR\\
        \hline
         Nuclei & 95.22$\pm$0.92 & 46.27$\pm$0.74 & 87.28$\pm$1.02 & 40.80$\pm$1.20  \\
         Vessels & \textbf{96.54$\pm$0.62}& \textbf{47.93$\pm$0.67} & \textbf{88.79$\pm$1.07 } & \textbf{41.95$\pm$0.96 }\\
        \hline
        \end{tabular}}
    \label{tab:ablation_modality}
\end{table}

\vspace{-3mm}
\paragraph{Impact of Different Modules in DAMM-Diffusion.} We conduct the experiments to verify the effectiveness of the proposed MMFM, UAFM and DAMMP modules, and summarize the results in \cref{tab:ablation_component}. As shown in the first three rows of \cref{tab:ablation_component}, the MMFM contributes to a 1.29\% SSIM and 1.32dB PSNR improvement respectively, and the UAFM improves the results by 1.76\% SSIM and 1.46dB PSNR. This benefits from the fact that both the MMFM and UAFM can fuse the multiple modalities effectively. Moreover, the performance will be further improved (shown in the 4-$th$ row of \cref{tab:ablation_component}) if we integrate the MMFM and UAFM together. Finally, we also observe that the DAMMP module is effective in generating the distribution of NPs (shown in the last row of \cref{tab:ablation_component}) since it can make a better choice between the uni-modal and multi-modal prediction results. 

\paragraph{Comparisons of UAFM with Cross-attention.}
We compare the generation performance of UAFM that considers the uncertainty in cross-attention with the traditional cross-attention module. The results shown in \cref{tab:cross-attention} indicate that UAFM performs better than its competitors since UAFM is helpful in measuring the divergence among different modalities to choose whether to use the uni-modal or multi-modal branch for predicting NPs distribution.

\vspace{-1mm}
\begin{table}[htbp] 
    \centering
    \tiny
    \caption{Ablation studies for each component in DAMM-Diffusion on internal validation set.}
    \resizebox{\linewidth}{!}{
    \scalebox{0.9}{
       \begin{tabular}{ccc| c c} 
       \hline
        MMFM& UAFM & DAMMP & SSIM \% & PSNR\\
        \hline
         & & & 92.92$\pm$1.17 & 43.96$\pm$1.03 \\
         $\checkmark$ & & & 94.21$\pm$0.72 & 45.28$\pm$0.73\\
          & $\checkmark$ & & 94.68$\pm$0.64 & 45.42$\pm$0.58 \\
         $\checkmark$& $\checkmark$& & 95.52$\pm$0.69 &  46.65$\pm$0.66 \\
         $\checkmark$& $\checkmark$& $\checkmark$& \textbf{96.54$\pm$0.62} & \textbf{47.93$\pm$0.67}\\
        \hline
        \end{tabular}}
    }
    \label{tab:ablation_component}
    \vspace{-3mm}
\end{table}

\begin{table}[htbp] 
    \centering
    \caption{Comparison with cross-attention on both internal and external validation set.}
    \resizebox{\linewidth}{!}{
       \begin{tabular}{c|cc| c c} 
       \hline
       {\multirow{2}{*}{Variants}} &\multicolumn{2}{c|}{Internal validation}&\multicolumn{2}{c}{External validation} \\
        & SSIM \% & PSNR & SSIM \% & PSNR\\
        \hline
         Cross-attention& 95.16$\pm$0.72 & 46.04$\pm$0.62 & 87.23$\pm$1.32 & 40.92$\pm$1.26  \\
         Ours & \textbf{96.54$\pm$0.62} & \textbf{47.93$\pm$0.67} & \textbf{88.79$\pm$1.07}& \textbf{41.95$\pm$0.96}\\
        \hline
        \end{tabular}}
    \label{tab:cross-attention}
\end{table}

\begin{table}[htbp] 
    \centering
    \tiny 
    \caption{Ablation studies for DFL on internal validation set.}
    \resizebox{\linewidth}{!}{
        \scalebox{0.9}{
       \begin{tabular}{c|cc| c c} 
       \hline
       {\multirow{2}{*}{$\gamma$}} &\multicolumn{2}{c|}{without $\mathcal{L}_{dfl}$}&\multicolumn{2}{c}{with $\mathcal{L}_{dfl}$} \\
        & SSIM \% & PSNR & SSIM \% & PSNR\\
        \hline
         0.3 & 94.25$\pm$0.79 & 44.36$\pm$0.85& 95.92$\pm$0.64 & 47.16$\pm$0.72  \\
         0.5 & 96.04$\pm$0.73 & 47.36$\pm$0.59 & \textbf{96.54$\pm$0.62}& \textbf{47.93$\pm$0.67}\\
         0.7 & 94.94$\pm$1.09 & 45.36$\pm$0.92 & 96.13$\pm$0.72 & 47.31$\pm$0.69 \\
        \hline
        \end{tabular}}
        }
    \label{tab:dfl}
    \vspace{-3mm}
\end{table}

\paragraph{Impact of Divergence Feedback Loss (DFL).} To further discuss the importance of the proposed DFL, we conduct experiments with different settings of $\gamma$, both with and without the use of DFL. The results are shown in \cref{tab:dfl}. We observe that the results with DFL are better than those without it when they are under the same value of $\gamma$. This benefits from the fact that DFL can adaptively adjust the uncertainty map according to the current results of both the uni-modal prediction and multi-modal prediction. Furthermore, the results in \cref{tab:dfl} also demonstrates that the choice of $\gamma$ significantly influences the results without DFL, since the low value of $\gamma$ encourages the adoption of the results from uni-modal branch while a high value of $\gamma$ always encourages fusion from the multi-modal data. Moreover, as shown in the right part of  \cref{tab:dfl}, the parameter $\gamma$ is in-sensitive to the DFL since it can give feedback to the divergence value to fit the comparison between uni-modal and multi-modal $L_1$ distance, which has been discussed before.

\vspace{-2mm}
\section{Conclusion}
\label{sec:conclu}
In this paper, we propose DAMM-Diffusion which can consider both uni-modal and multi-modal branches to perform the reverse step of the diffusion model in a unified network. The experimental results on predicting the NPs distribution pixels-to-pixels verify the advantages of DAMM-Diffusion in comparison with the existing studies.

\section*{Acknowledgments}
This work is supported by the National Natural Science Foundation of China (Nos.62136004,62272226,82372019).
{
    \small
    \bibliographystyle{ieeenat_fullname}
    \bibliography{main}
}

\clearpage
\setcounter{page}{1}
\maketitlesupplementary

\section{Method}
\subsection{Preliminaries of Diffusion Models}
Diffusion Models are consisted of two processes: the forward process and the reverse process. The forward process progressively perturbs $x_0$ to a latent variable by adding noise sampling from isotropic Gaussian distribution. Mathematically, a $T$-step forward process can be formulated as the following Markovian chain: 
\begin{equation}
q(x_1,...,x_T|x_0)=\prod_{t=1}^{T}q(x_t|x_{t-1}),
\label{eq:forward}
\end{equation}
where $q(x_t|x_{t-1})=\mathcal{N}(x_t;\sqrt{1-\beta_{t}}x_{t-1}, \beta_{t}\mathbf{I})$ is a normal distribution whose mean value is $\sqrt{1-\beta_{t}}x_{t-1}$ and the deviation is $\beta_t\mathbf{I}$. Here, $\beta_t$ is the variance schedule across diffusion steps. The latent variable $x_{T}\sim\mathcal{N}(0,\mathbf{I})$  when $T{\rightarrow}\infty$.

The reverse process can be viewed as a corresponding denoise process to recover $x_0$ from the latent variable $x_T$, which can be parameterized as: 
\begin{equation}
p_{\theta}(x_0,...,x_{T-1}|x_{T})=\prod_{t=1}^{T}p_{\theta}(x_{t-1}|x_t),
\end{equation}
where $p_{\theta}(x_{t-1}|x_t)$ is represented as the approximate Gaussian such that $p_{\theta}(x_{t-1}|x_t)=\mathcal{N}(x_{t-1};\mu_{\theta}(x_t,t),\Sigma_{\theta}(x_t,t))$, $\mu_{\theta}(x_t,t)$ and $\Sigma_{\theta}(x_t,t))$ are the mean and variance which can be estimated by $\theta$. In practice, the variance is set to untrained time dependent constants i.e., $\Sigma_{\theta}(x_t,t)=\beta_t\mathbf{I}$.

The objective of the Diffusion Model is to maximize the Evidence Lower Bound (ELBO) of the joint distribution of forward process, which can be simplified as: 
\begin{equation}
\mathbb{E}_{x_0,t,\epsilon}||\epsilon-\epsilon_{\theta}(x_t,t)||^2_2,
\end{equation}
where $\epsilon\sim \mathcal{N}(0, I)$ is the Gaussian noise added in $x_t$ and $\theta$ represents the parameter of a neural network.

Conditional Diffusion Models (CDMs) aim to implement controllable diffusion with condition $y$ for jointly training, and the objective can be modified as: 
\begin{equation}
\mathbb{E}_{x_0,y,t,\epsilon}||\epsilon-\epsilon_{\theta}(x_t,y,t)||^2_2.
\end{equation}
For the image-to-image translation task, the condition $y$ is the image in the source domain. 

Latent Diffusion Models (LDMs) ~\cite{rombach2022high} operate the forward and reverse processes in a latent space rather than the original pixel space which help focus on the important semantic information of the data while mitigating the need for redundant and intensive computations.

\subsection{Computational Complexity} 
Here, we focus on analyzing the computational complexity of the MMFM and UAFM modules.
MMFM is consisted of two parts i.e., spatial attention and channel attention. The computational complexity for the spatial attention with the input feature $v\in R^{C\times H \times W}$  in \cref{eq:mmfm_1} is $2O(C^2\times K^2\times H\times W)+4O(C\times H\times W)$, where the computational complexity for each convolution, normalization and activation operations are $O(C^2\times K^2\times H\times W)$, $O(C\times H\times W)$ and $O(C\times H\times W)$, respectively. Similarly, the computational complexity of the channel attention with the concatenated input $f_{sp}\in R^{2C\times H\times W}$ in \cref{eq:mmfm_2} is $O(2C\times H\times W)+O(\frac{2C}{r}\times 2C)+O(2C\times \frac{2C}{r})$, where the complexity of AvgPool and linear operations are $O(2C\times H\times W)$ and $O(\frac{2C}{r}\times 2C)+O(2C\times \frac{2C}{r})$, respectively. \textbf{In summary, the total computational complexity of the MMFM module is} $2O(C^2\times K^2\times H\times W)+4O(C\times H\times W)+O(2C\times H\times W)+O(\frac{2C}{r}\times 2C)+O(2C\times \frac{2C}{r})$. On the other hand, UAFM mainly involves the calculation of uncertainty-aware cross-attention (shown in \cref{eq:uaca}). Thus, \textbf{the total complexity for UAFM module is} $2O(H^{2}\times W^{2}\times d)+O(H^{2}\times W^{2})$. 


\section{Additional Experiments}

\subsection{Additional Results of Uni-modal Methods}
We present additional results for each individual modality (i.e., nuclei and vessels) of uni-modal methods in \cref{suptab:uni-modal}. We can observe that our method consistently outperforms the uni-modal methods for both nuclei and vessels.
The results also support the finding that vessels are more beneficial for predicting NPs.
\begin{table*}[htpb]
    \centering
    \caption{\centering{Performance comparisons with uni-modal methods on internal validation set. The symbol \textsuperscript{*} indicates significant improvement ($p<0.05$).}}
    \vspace{-0.6em}
    \resizebox{0.65\linewidth}{!}{
    \begin{tabular}{c|cc cc}
        \hline
        Methods & SSIM \% (nuclei) & SSIM \% (vessels) & PSNR (nuclei) & PSNR (vessels)\\
        \hline
         Cyclegan& 74.87$\pm$3.64 $^{*}$ & 84.07$\pm$2.67 $^{*}$ & 28.12$\pm$2.46 $^{*}$ & 36.96$\pm$2.34 $^{*}$  \\
         Pix2pix & 76.02$\pm$3.24 $^{*}$ & 87.81$\pm$2.15 $^{*}$ & 31.27$\pm$2.12 $^{*}$  & 38.97$\pm$2.70 $^{*}$ \\
         LDM& 78.23$\pm$1.02 $^{*}$ & 92.97$\pm$0.65  $^{*}$ & 32.57$\pm$1.17 $^{*}$ & 43.72$\pm$0.62  $^{*}$\\
         BBDM & 79.05$\pm$1.31 $^{*}$ & 93.01$\pm$0.81  $^{*}$ & 32.34$\pm$0.96 $^{*}$  & 43.96$\pm$0.75 $^{*}$ \\ 
         Ours & \multicolumn{2}{c}{\textbf{96.54$\pm$0.62}} & \multicolumn{2}{c}{\textbf{47.93$\pm$0.67}} \\
        \hline
    \end{tabular}} \\
    \label{suptab:uni-modal}
\end{table*}

\subsection{Settings of Hyperparameters.}
\begin{figure}[htbp]
  \centering
  \includegraphics[width=0.8\linewidth]{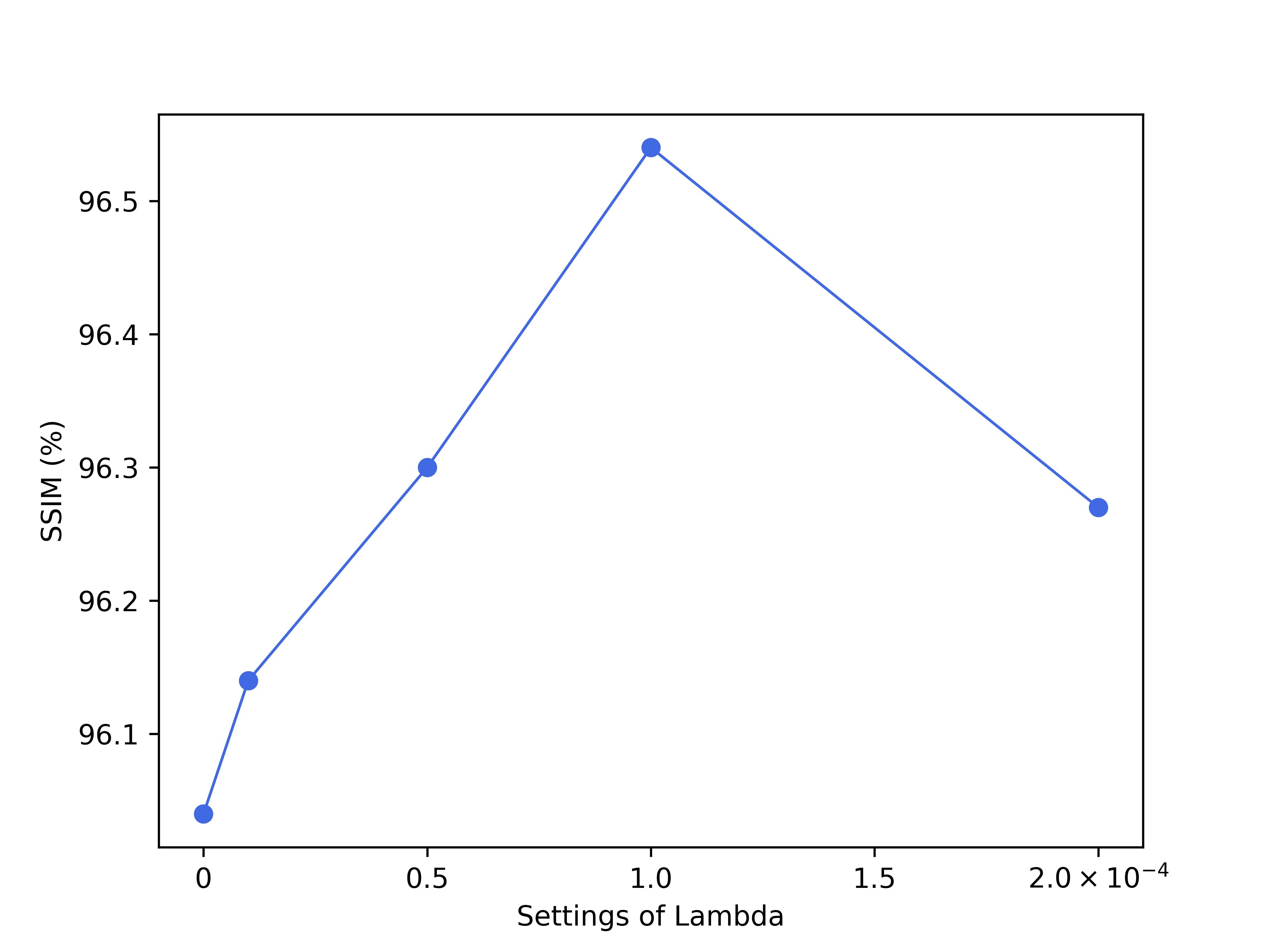}
  \caption{The effect of hyperparameter $\lambda$.}
  \label{sup_fig:lambda}
\end{figure}

\begin{figure}[htbp]
  \centering
  \includegraphics[width=0.8\linewidth]{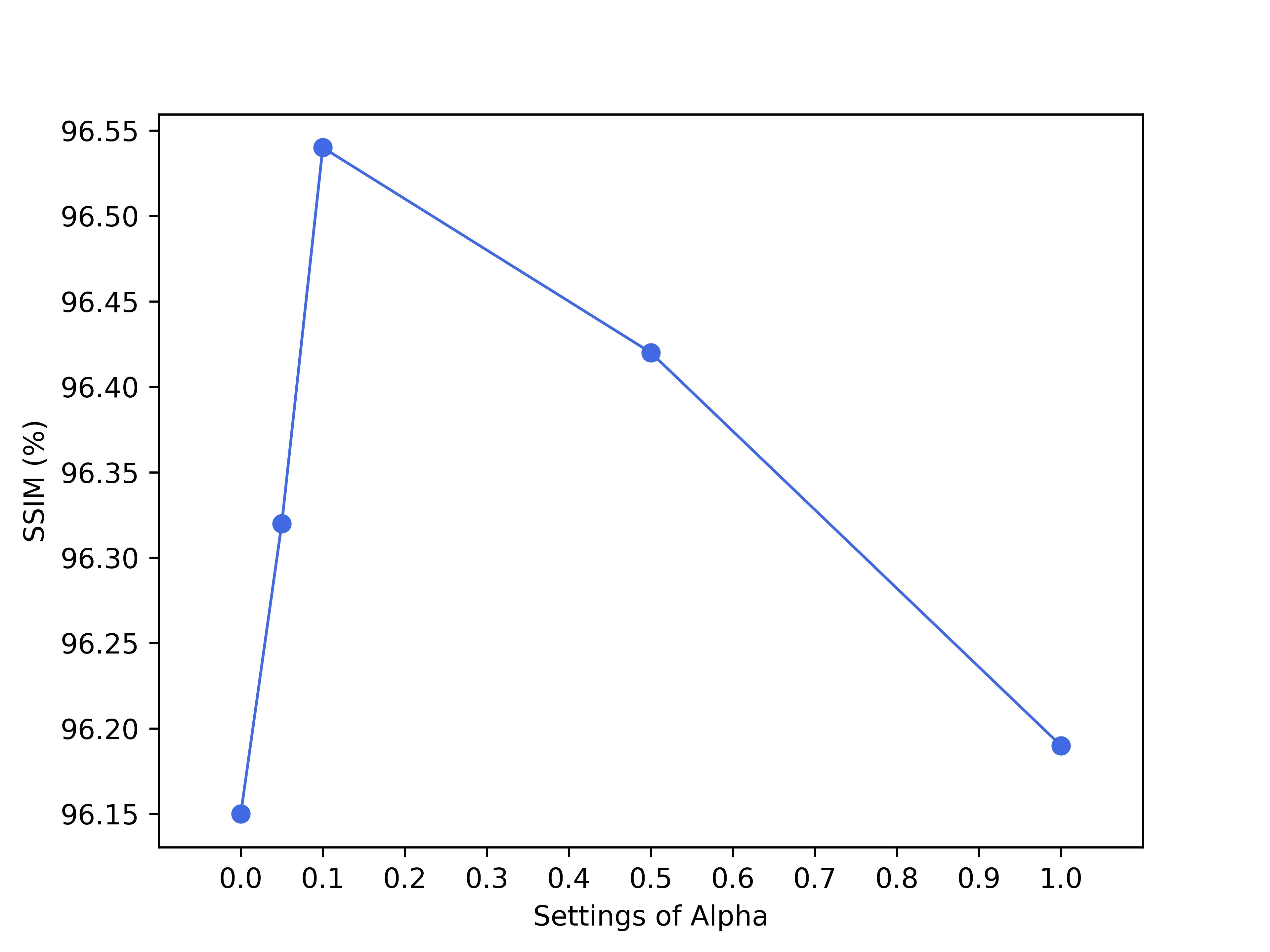}
  \caption{The effect of hyperparameter $\alpha$.}
  \label{sup_fig:alpha}
\end{figure}

We conduct studies about the hyperparameters of $\lambda$ in \cref{eq:dflloss} and $\alpha$ in \cref{eq:totalloss} on the internal validation, with results in \cref{sup_fig:lambda} and \cref{sup_fig:alpha}. Based on outcomes across different datasets, we find that $\lambda$ achieves optimal performance at 1e-4, while setting $\alpha$ to 0.1 is more beneficial for the results.

\subsection{Different Types of Datasets and Tasks}
\label{supsec:brats}
\paragraph{Dataset.}
We further validate the effectiveness of DAMM-Diffusion on the brain image synthesis task.
Specifically, we test our DAMM-Diffusion on the Multi-modal Brain Tumor Segmentation Challenge 2018 (BRATS) dataset~\cite{menze2014multimodal}. The BRATS dataset consists of 285 patients with the multi-modal MRI scans including different imaging modalities: $T_1$, $T_{2}$ and $FLAIR$. These scans were acquired using various clinical protocols and scanners from 19 different institutions, ensuring a diverse and comprehensive dataset. Each modality volume has a size of 240 $\times$ 240 $\times$ 155 voxels. 
In this study, we automatically select 2D axial-plane slices, crop a central 200 $\times$ 200 region from each and then resize it to 256 $\times$ 256. Additionally, we randomly split the 285 subjects to 80\% for training and 20\% for testing.

\paragraph{Results of Different Modalities on Uni-modal Branch.}
We compare the performance of DAMM-Diffusion when using different input modalities in the uni-modal branch on the BRATS dataset.
As shown in \cref{sup_tab:channel}, the results indicate that the choice of different input images do not significantly impact the final performance on the BRATS dataset. This may be due to the fact that each modality in the multi-modal brain image synthesis effectively contributes to the overall outcomes.

\renewcommand{\tabcolsep}{2pt}
\renewcommand{\arraystretch}{1.20}
\begin{table}[htbp]
\centering
\resizebox{1.\columnwidth}{!}{%
\begin{tabular}{ccccccc}
\hline
Task & \multicolumn{2}{c}{\ToneTtwoFlair}  & \multicolumn{2}{c}{\ToneFlairTtwo} & \multicolumn{2}{c}{\TtwoFlairTone}  \\ \cline{1-7} 
Input & T\textsubscript{1}& T\textsubscript{2}& T\textsubscript{1}& FLAIR& T\textsubscript{2}& FLAIR   \\ \hline
\multirow{2}{*}{SSIM}& 
88.90& 89.20 & 92.37& 92.86      & 92.68 &  92.27\\
&$\pm$6.25 &$\pm$5.65 & $\pm$3.66& $\pm$3.60&   $\pm$4.22&  $\pm$4.42\\ \hline
\multirow{2}{*}{PSNR}&
24.13& 	24.23& 	25.07& 	25.25& 	25.61&	25.29  \\
&$\pm$3.38&  $\pm$3.16&  $\pm$3.74&   
$\pm$3.48&   $\pm$2.49&   $\pm$2.60\\ \hline

\end{tabular}}
\caption{
Effects of choosing different types of images in the
uni-modal branch on the BRATS Dataset.}
\label{sup_tab:channel}
\vspace{-1ex}
\end{table}

\begin{figure*}[htbp]
  \centering
  \includegraphics[width=\linewidth]{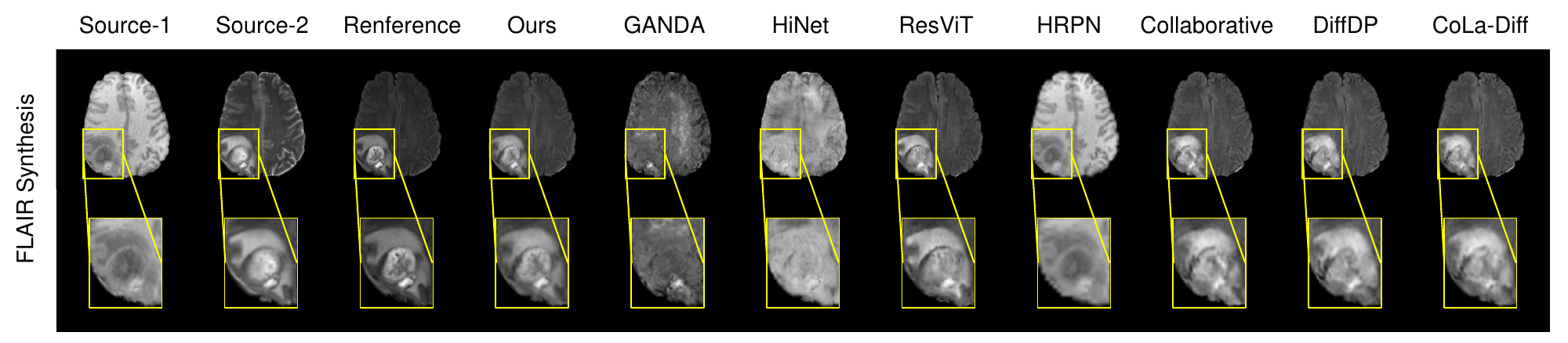}
  \caption{Visualization results of benchmark methods and DAMM-Diffusion on the BRATS dataset for the representative many-to-one synthesis task: \ToneTtwoFlair.}
  \label{sup_fig:brats_flair}
\end{figure*}

\begin{figure*}[htbp]
  \centering
  \includegraphics[width=\linewidth]{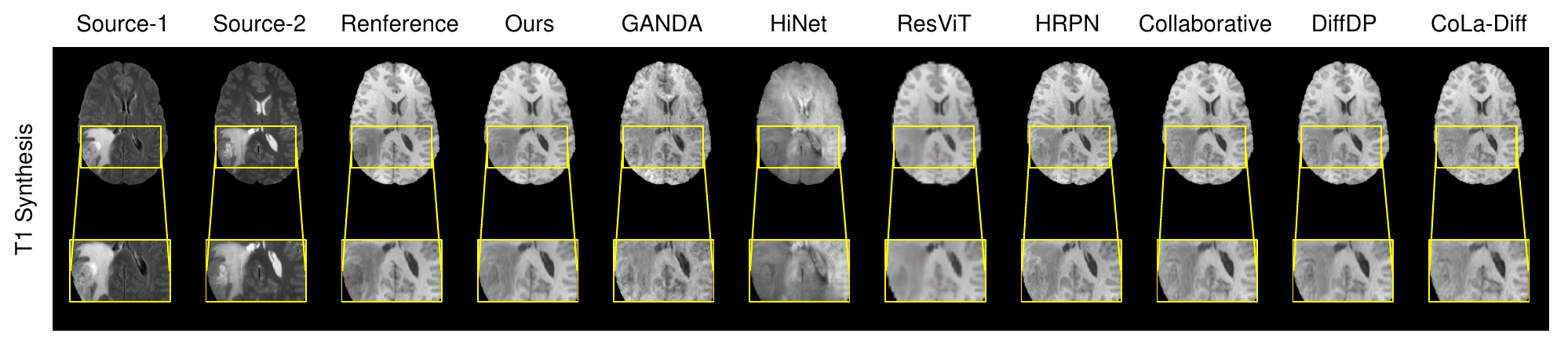}
  \caption{Visualization results of benchmark methods and DAMM-Diffusion on the BRATS dataset for the representative many-to-one synthesis task: \TtwoFlairTone.}
  \label{sup_fig:brats_t1}
\end{figure*}

\begin{figure*}[htbp]
  \centering
  \includegraphics[width=\linewidth]{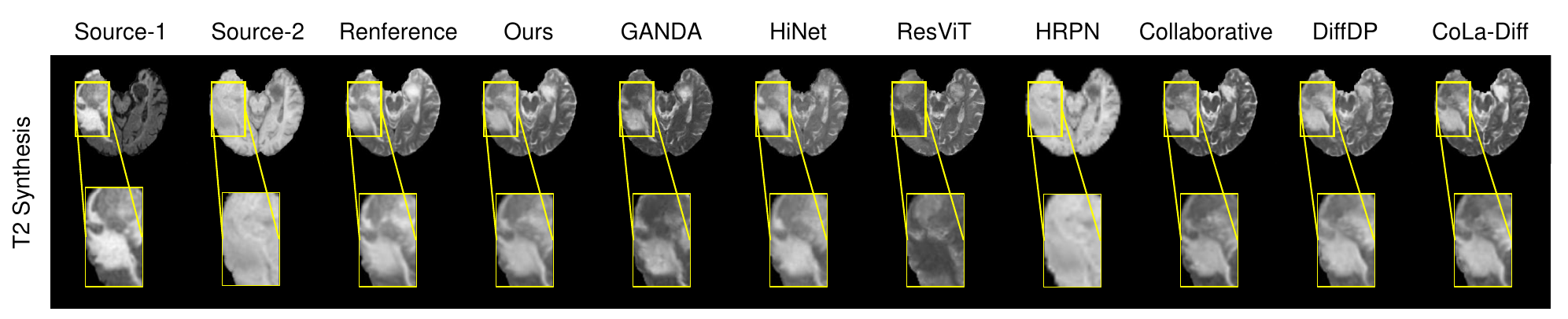}
  \caption{Visualization results of benchmark methods and DAMM-Diffusion on the BRATS dataset for the representative many-to-one synthesis task: \ToneFlairTtwo.}
  \label{sup_fig:brats_t2}
\end{figure*}

\paragraph{Qualitative Results.}
We present the representative target images for \ToneTtwoFlair~, \TtwoFlairTone~ and \ToneFlairTtwo~ in \cref{sup_fig:brats_flair}, \cref{sup_fig:brats_t1} and \cref{sup_fig:brats_t2}, respectively. 
Compared to the baseline methods, our approach generates target images with significantly reduced artifacts and enhanced clarity in tissue depiction. 
As shown in \cref{sup_fig:brats_flair}, DAMM-Diffusion can accurately capture brain lesions and provide the details of pathological regions, while the other methods fail to achieve. 
These results demonstrate the superiority of DAMM-Diffusion in generating the reliable medical images.

\subsection{Additional Visualization Analysis}

\begin{figure*}[htbp]
  \centering
  \includegraphics[width=0.95\linewidth]{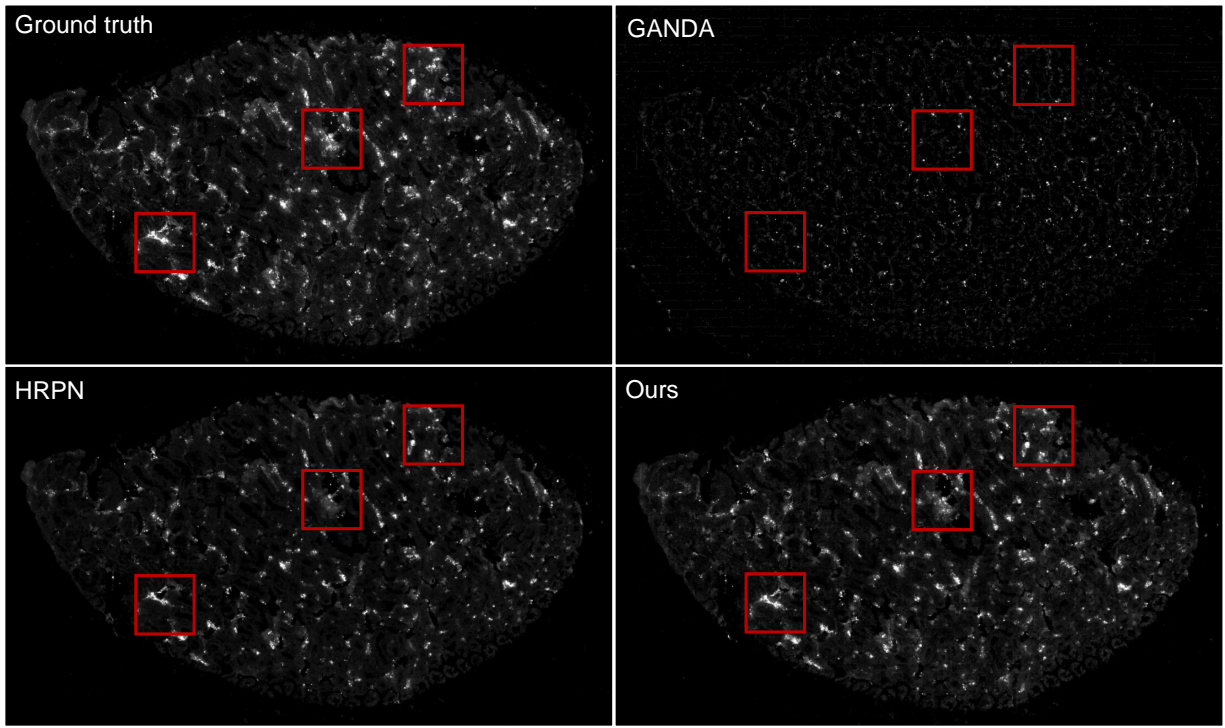}
  \caption{Qualitative comparison between the proposed method and the previous methods for NPs distribution prediction in a whole-slide image.}
  \label{sup_fig:fig_ws}
  \vspace{-0.5em}
\end{figure*}

\begin{figure*}[htbp]
  \centering
  \includegraphics[width=\linewidth]{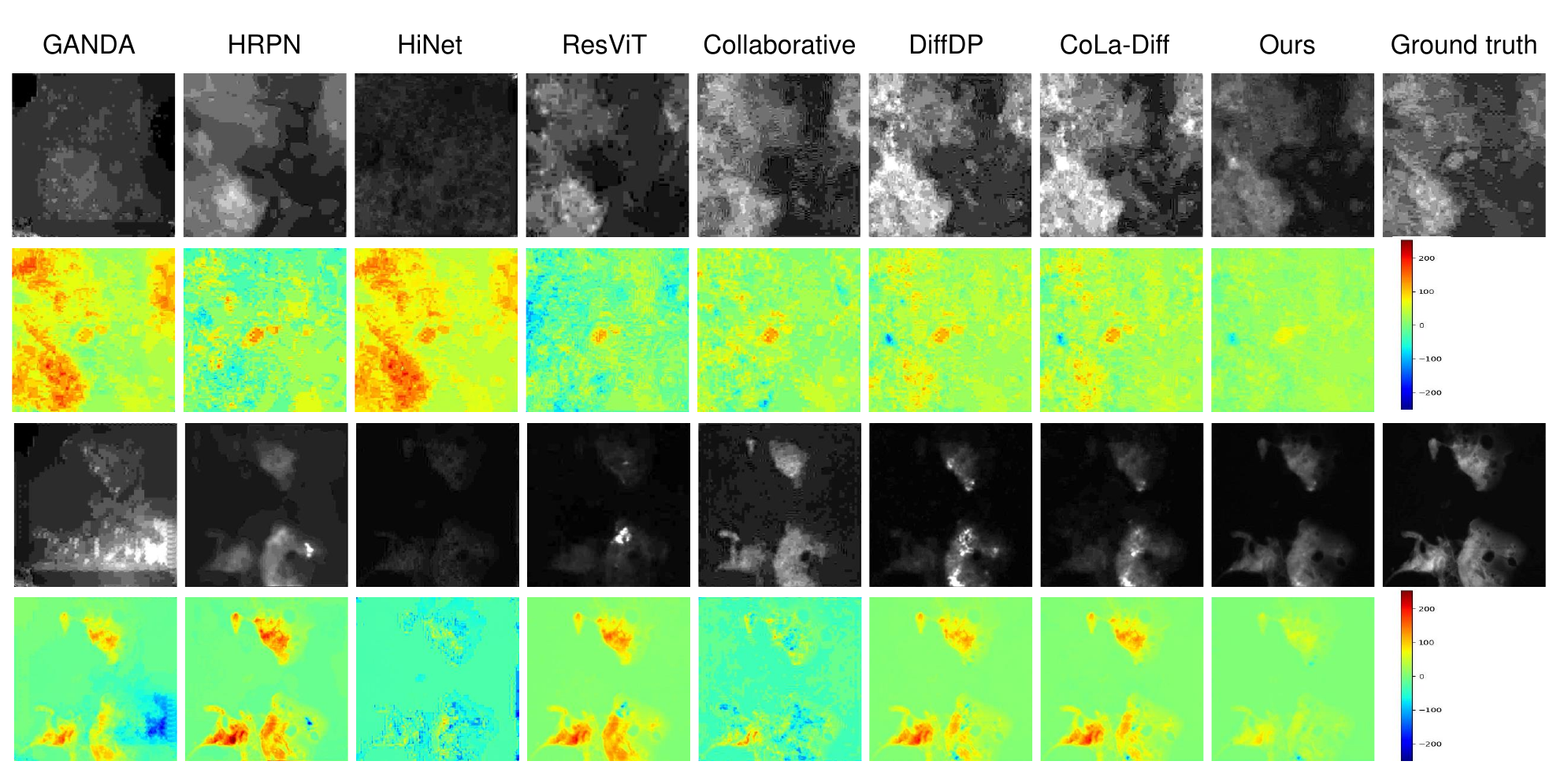}
  \caption{Visualization of generated NPs distribution (1st and 3th rows) and corresponding difference maps (2nd and 4th rows) at patch level.}
  \label{sup_fig:fig_patch}
\end{figure*}

We provide more visualization results on the NPs distribution prediction task, including the generated whole-slide and patch-level images in \cref{sup_fig:fig_ws} and \cref{sup_fig:fig_patch}, respectively.

\end{document}